\pdfoutput=1  

\documentclass{article}
\usepackage{ijcai26}

\usepackage{times}
\usepackage{soul}
\usepackage{url}
\usepackage[hidelinks]{hyperref}
\usepackage[utf8]{inputenc}
\usepackage[small]{caption}
\usepackage{graphicx}
\usepackage{amsmath}
\usepackage{amssymb}
\usepackage{amsthm}
\usepackage{booktabs}
\usepackage{multirow}
\usepackage{xcolor}
\usepackage{tikz}
\usetikzlibrary{positioning,arrows.meta,fit,backgrounds}
\definecolor{lightblue}{RGB}{189,215,238}
\definecolor{accentblue}{RGB}{68,114,196}
\definecolor{alertred}{RGB}{192,0,0}
\usepackage[switch]{lineno}

\graphicspath{{figures/}}

\newcommand{\MAVRP}{MAVRP}
\newcommand{\Att}{\textsc{Att}}
\newcommand{\Unimp}{\textsc{Unimp}}
\newcommand{\Moe}{\textsc{UnimpMoe}}
\newcommand{\EtoE}{\textsc{Hard-Mask}}
\newcommand{\CA}{\textsc{Recourse}}

\title{Two Black Boxes, One Solver: Encoder Probing and Decoder Attribution for Neural Multi-Attribute VRP under Hard-Mask and Recourse Decoders}

\author{
    Sohaib Afifi
    \affiliations
    Univ. Artois, UR 3926, Laboratoire de Génie Informatique et d’Automatique de l'Artois (LGI2A),\\F-62400 Béthune, France
    \emails
    sohaib.afifi@univ-artois.fr
}

\begin{document}
\maketitle

\begin{abstract}
Neural autoregressive solvers for the Multi-Attribute Vehicle Routing Problem (MAVRP) reach competitive cost but offer no per-step justification, a problem when dispatchers must validate, accept, or compare them. We open two complementary black boxes in one protocol. On the \emph{encoder} side, linear probes, spontaneous-organization metrics, rank-based richness measures, and discovered-direction analyses with intervention validation characterize how the latent represents constraint families at the graph, node, and edge level. On the \emph{decoder} side, three attribution methods (gradient, integrated gradients, DeepLIFT) feed three reading angles: abductive, contrastive against the best feasible alternative, and counterfactual (smallest input change that switches the action or restores feasibility). Explanations are scored on fidelity, concentration, stability, sanity, and actionability. Across six variants combining three encoders ($\Att$ baseline, $\Unimp$, $\Moe$) with two decoders ($\EtoE$, $\CA$), we find that graph inductive bias improves both representational predictability and decoder sanity, that the Mixture-of-Experts encoder represents constraints in a distributed rather than axis-aligned way, and that the $\CA$ training regime, not merely its softer mask, produces policies that represent infeasibility usefully, exposing make-feasible counterfactuals that $\EtoE$ policies fail to produce even when fed infeasible alternatives externally.
\end{abstract}

\section{Introduction}

The Multi-Attribute Vehicle Routing Problem (\MAVRP) packs several VRP variants into a single combinatorial problem: a single instance can simultaneously activate capacity, time-window, distance-limit, backhaul, and open-route constraints~\cite{kool2019attention,berto2024rl4co}. Neural autoregressive solvers built on attention~\cite{vaswani2017attention,kool2019attention} or graph encoders such as UniMP~\cite{shi2021unimp}, optionally with conditional compute through a Mixture-of-Experts (MoE) layer~\cite{shazeer2017outrageously}, now match strong heuristics on these benchmarks, including graph-based unified models built specifically for \MAVRP~\cite{jari2025gunirouting}. They remain, however, opaque: at every decoding step the policy emits an action with no exposed justification of which inputs mattered, which constraints dominated, which alternative was close, or what minimal change would have flipped the decision.

This opacity is an operational liability rather than a purely epistemic one. Logistics dispatchers need (i) \emph{diagnosis} of anomalous routes, (ii) \emph{validation} against explicit business rules, (iii) \emph{acceptability} of the solver's recommendations, and (iv) \emph{comparability} between candidate solver families. These four needs cut across both representation and decision: an encoder may legibly encode a constraint that the decoder ignores, and a decoder may attribute mass to a feature whose representation is degenerate. We therefore argue that neural CO solvers require a \emph{joint} XAI account of both stages, evaluated on the same checkpoints with a shared grid.

We make four contributions. First, we propose a two-pillar protocol that probes the encoder for predictability, spontaneous organization, richness, and discovered directions, while attributing decoder decisions through three complementary reading angles. Second, we introduce a constraint-family taxonomy that aggregates raw input gradients into operations-readable groups (capacity/demand, geometry, time windows, route-recourse). Third, we apply the protocol to six \MAVRP\ solver variants combining three encoder families ($\Att$ baseline, $\Unimp$, $\Moe$) with two decoder variants ($\EtoE$ and $\CA$) and report under five XAI criteria. We deliberately compare a strong-mask decoder ($\EtoE$, feasibility-by-construction) against a weak-mask decoder ($\CA$, feasibility-by-learning) because the gap is itself the contribution: the recourse decoder is the only one in which ``why was this alternative not chosen?'' has a non-trivial answer, and where \emph{make-feasible} counterfactuals can exist at all. Fourth, we expose phenomena invisible to inference cost alone: distributed coding in $\Moe$, sanity gap between $\Att$ and the graph encoders, and a representational gap between $\EtoE$ and $\CA$ policies that make-feasible counterfactuals reveal even after we control for the mask.

\section{Setup}

\paragraph{Problem.} An instance of \MAVRP\ is a directed graph $G=(V,E)$ with one depot and $n$ customers. Each customer carries linehaul/backhaul demands, a service time, and a time window; each instance carries a vehicle capacity, an optional distance limit, and a flag for open routes. Following the unified parameterization of~\cite{berto2024rl4co} we sample instances such that any subset of constraint families may be active, so the same encoder--decoder pair must handle every variant. We use $n=50$ customers throughout.

\paragraph{Solver family.} The policy $\pi_\theta$ first encodes the whole instance into per-node embeddings $h \in \mathbb{R}^{n \times d}$, then autoregressively decodes a permutation by attending over the masked candidate set. We compare three encoders: $\Att$, a Transformer-style attention encoder used as the \emph{baseline} architecture; $\Unimp$, a graph encoder with structural inductive bias~\cite{shi2021unimp,jari2025gunirouting}; $\Moe$, the same backbone augmented with a Mixture-of-Experts layer for conditional compute~\cite{shazeer2017outrageously}.

\paragraph{Decoder variant.} We pair every encoder with two decoder variants that differ only in how feasibility is delivered. The $\EtoE$ decoder enforces every active constraint (capacity, time windows, distance, backhaul, open-route) directly in the action mask: only feasible candidates ever reach the softmax, so feasibility is guaranteed by construction. The $\CA$ decoder masks only already-visited customers and leaves the policy itself to learn the constraints; infeasible selections are allowed during training and pay a recourse cost in the reward, and the decoder is given per-step state features (remaining capacity, time slack, recourse cost) so it can become aware of constraint slack at decoding time. Both share the same attention-masked autoregressive backbone, so the comparison isolates the effect of how feasibility is delivered. The six (encoder $\times$ decoder) checkpoints are the unit of comparison throughout the paper.

\begin{figure*}[t]
\centering
\begin{tikzpicture}[
    x=1mm, y=1mm,
    >=Latex,
    box/.style       ={draw, rounded corners=1.5pt, minimum height=8mm, minimum width=22mm, font=\small, align=center, inner sep=2pt},
    sub/.style       ={font=\scriptsize, align=center, text=black!70},
    pillar/.style    ={draw=#1, dashed, rounded corners=2.5pt, fill=#1!8, line width=0.5pt, inner sep=0pt},
    pilltitle/.style ={font=\footnotesize\bfseries, text=#1, align=center},
    pillitem/.style  ={font=\scriptsize, align=left, anchor=west, inner sep=0pt},
    flow/.style      ={-{Latex[length=2mm,width=1.5mm]}, thick, black!70},
    loop/.style      ={-{Latex[length=2mm,width=1.5mm]}, thick, dashed, gray}
]
\node[box, fill=lightblue!25]                              (inst)   at (  0, 0) {Instance};
\node[box, fill=lightblue!40, right=10mm of inst]          (enc)                 {Encoder};
\node[box, fill=lightblue!55, right=10mm of enc]           (lat)                 {Latent};
\node[box, fill=lightblue!75, right=10mm of lat]           (dec)                 {Decoder};
\node[box, fill=accentblue!90, text=white, right=10mm of dec] (act)              {action $a_t$};

\node[sub, below=0.5mm of inst.south] {$\{x_i\}_{i=1}^n,\ g$};
\node[sub, below=0.5mm of enc.south]  {\Att\,$\mid$\,\Unimp\,$\mid$\,\Moe};
\node[sub, below=0.5mm of lat.south]  {$h \in \mathbb{R}^{n\times d}$};
\node[sub, below=0.5mm of dec.south]  {$\EtoE\,\mid\,\CA$};
\node[sub, below=0.5mm of act.south]  {logits $\to$ softmax};

\draw[flow] (inst) -- (enc);
\draw[flow] (enc)  -- (lat);
\draw[flow] (lat)  -- (dec);
\draw[flow] (dec)  -- (act);

\draw[loop]
    (act.north) -- ++(0,5)
    node[font=\scriptsize, midway, right=1mm, text=gray]{state $\&$ mask update}
    -| (dec.north);

\path (enc.south) -- node[midway] (p1c) {} (lat.south);
\node[pillar=accentblue, minimum width=72mm, minimum height=24mm,
      anchor=north] (p1) at ([yshift=-16mm]p1c) {};
\node[pilltitle=accentblue, anchor=north, yshift=-1.2mm] (p1t) at (p1.north)
      {\textsc{Pillar 1}: encoder probes  (\S 4)};
\node[font=\scriptsize, align=left, anchor=north west,
      text width=66mm, inner sep=0pt]
      at ([xshift=3mm, yshift=-1.4mm]p1t.south west) {%
        $\bullet$\ supervised probes (AUC, F1)\\
        $\bullet$\ axis-unique vs.\ subspace\\
        $\bullet$\ NMI / silhouette / eff.\ rank\\
        $\bullet$\ PCA / ICA $+$ intervention};
\draw[->, dashed, accentblue, thick] (lat.south) -- (p1.north);

\path (dec.south) -- node[midway] (p2c) {} (act.south);
\node[pillar=alertred, minimum width=72mm, minimum height=24mm,
      anchor=north] (p2) at ([yshift=-16mm]p2c) {};
\node[pilltitle=alertred, anchor=north, yshift=-1.2mm] (p2t) at (p2.north)
      {\textsc{Pillar 2}: decoder attribution  (\S 5)};
\node[font=\scriptsize, align=left, anchor=north west,
      text width=66mm, inner sep=0pt]
      at ([xshift=3mm, yshift=-1.4mm]p2t.south west) {%
        $\bullet$\ abductive: top-$k$ on $\partial u_{t,a_t}$\\
        $\bullet$\ contrastive: $\partial(u_{t,a_t}{-}u_{t,a'_t})$\\
        $\bullet$\ counterfactual: switch / make-feas.\\
        $\bullet$\ scored on 5 criteria};
\draw[->, dashed, alertred, thick] (dec.south) -- (p2.north);

\end{tikzpicture}
\caption{Solver pipeline (top row) and the two XAI pillars. The pipeline encodes the instance into a per-node latent $h$ and autoregressively decodes a permutation; the dashed gray arrow is the action loop: at every step $t$ the chosen $a_t$ updates the decoder state and the feasibility mask before the next decision. \textsc{Pillar 1} probes the frozen latent (\S 4); \textsc{Pillar 2} attributes the per-step action under three reading angles, scored on five criteria (\S 5). Both pillars share the same six (encoder family $\times$ decoder variant) checkpoints, so all comparisons are made on the same trained policies.}
\label{fig:arch}
\end{figure*}
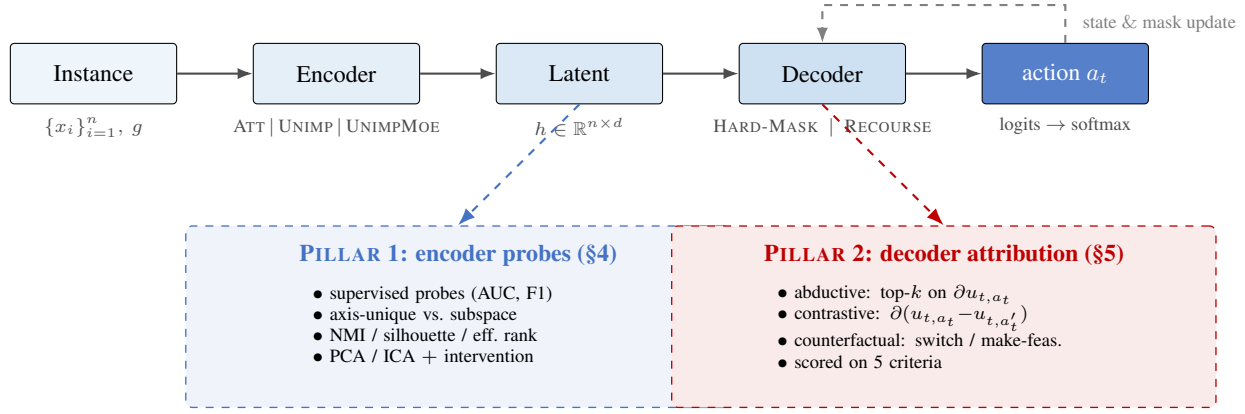

\section{Method}

We instrument both stages of the trained policy and score every output on a single five-criterion grid. The protocol is implemented as a batch runner that consumes a checkpoint and returns a structured report combining encoder probes, decoder attributions, and evaluation metrics. We run three seeds per (encoder $\times$ decoder $\times$ method) cell over $128$ instances.

\subsection{Encoder Probing}

We ask four questions of the frozen latent $h$, all answered at the graph, node, and edge level.

\paragraph{Predictability.} Can the active constraint families be linearly recovered from $h$? We train per-family linear probes~\cite{alain2016understanding} and report ROC-AUC for binary indicators and macro-F1 for multi-class targets. Two complementary unsupervised readouts are computed on PCA/ICA components of $h$: an \emph{axis-unique} score (best correlation with a single component) and a \emph{top-$k$ subspace} score (best correlation with a $k$-dimensional subspace). The gap between the two diagnoses distributed coding: information that lives across components rather than along a single axis.

\paragraph{Spontaneous organization.} Without labels, $k$-means on $h$ should partition instances by constraint family. We score the partition with silhouette~\cite{rousseeuw1987silhouettes}, NMI~\cite{vinh2010information}, and ARI~\cite{hubert1985comparing}, and visualise it with t-SNE~\cite{vandermaaten2008tsne}.

\paragraph{Richness.} The effective rank and stable rank of $h$~\cite{roy2007effective} detect collapse to a low-dimensional subspace and serve as denominators in the rank-vs-organization synthesis (Figure~\ref{fig:encoder}).

\paragraph{Discovered directions.} PCA and ICA decompositions of $h$ produce candidate latent axes; we then align each axis with a known reference bank of constraint indicators and concept descriptors. To validate the resulting alignments causally, we intervene along each direction in the latent and check whether the targeted concept moves in the predicted direction (\emph{intervention validation}); we also measure the across-seed stability of these directions.

\subsection{Decoder Attribution: Three Reading Angles}

At decoding step $t$ the policy emits logits $u_{t,a}$ over the masked candidate set; let $a_t = \arg\max_a u_{t,a}$ be the chosen action and $a'_t$ the best feasible alternative. We attribute every step with three input attribution methods (gradient, integrated gradients~\cite{sundararajan2017axiomatic}, and DeepLIFT~\cite{shrikumar2017learning}) using Captum~\cite{kokhlikyan2020captum}.

\paragraph{Abductive.} For each method we obtain an input-attribution tensor $g_{i,f}$ at step $t$, then aggregate it to per-node $s_i = \sum_f |g_{i,f}|$, per-feature $s_f = \tfrac{1}{N}\sum_i |g_{i,f}|$, and per-constraint-family $s_{\mathrm{fam}} = \sum_{f \in \mathrm{fam}} s_f$. The top-$k$ on each level answers \emph{``why this action?''}.

\paragraph{Contrastive.} We compute the margin $\Delta_t = u_{t,a_t} - u_{t,a'_t}$ and attribute $\partial \Delta_t / \partial x_{i,f}$. Features whose contrastive attribution is largest are those that \emph{depart} the chosen action from its closest feasible rival. The margin $|\Delta_t|$ also doubles as a confidence signal: a large value indicates a clear preference, a small value a tight call.

\paragraph{Counterfactual.} A directed search guided by the contrastive gradient looks for the smallest input perturbation that either (i) flips the chosen action (\emph{switch}) or (ii) restores feasibility for an infeasible alternative (\emph{make-feasible}). Both modes test \emph{actionability}: can a dispatcher reach a different decision through a small, plausible change to the instance? We specify the search fully rather than tune it. It runs only when a feasible alternative $a'_t$ exists and the margin $\Delta_t = u_{t,a_t} - u_{t,a'_t}$ is finite and positive, and ranges over a fixed dictionary of seven signed relaxations, each pushing a binding constraint the way that helps: node-level time-window end ($\uparrow$), service time ($\downarrow$), linehaul and backhaul demand ($\downarrow$), and instance-level vehicle capacity, distance limit and depot time-window end ($\uparrow$). For each candidate $c$ the first-order step $\delta_c = \Delta_t / |\partial \Delta_t / \partial x_c|$ estimates the change that closes the margin; we discard it if the gradient points the wrong way, if the move is inadmissible (e.g.\ a value below zero), or if its relative size $\delta_c / \max(|x_c|,1)$ exceeds a $5\times$ budget cap. The three smallest survivors are verified by applying the move (with a $5\%$ overshoot) and replaying one decode step, recording a switch if $a'_t$ becomes the $\arg\max$ or a make-feasible if the infeasible $a'_t$ turns feasible; we return the first success, else the smallest attempt. The search is thus deterministic with a fixed budget of three replays, so the rates in \S 5.7 are lower bounds a larger budget could only raise.

\subsection{Five-Criterion Evaluation}

Each criterion answers one operational question; we state what it measures and what a dispatcher reads from it. For each (model, method) we report:
\begin{itemize}
    \item \textbf{Fidelity}: \emph{do the highlighted inputs really drive the decision?} A deletion test perturbs the top-$k$ attributed nodes and measures the action flip rate (\textit{flip@$k$}) and log-probability drop (\textit{$\Delta$logp@$k$}); a faithful explanation flips the action when its own top nodes are removed.
    \item \textbf{Concentration}: \emph{focused or diffuse?} \textit{focus@$k$} is the mass on the top-$k$ features and \textit{clarity} is one minus the normalised entropy of the attribution, so a sharp story scores high and a smeared one low.
    \item \textbf{Stability}: \emph{would a differently seeded model tell the same story?} We re-run the attribution on independently seeded checkpoints of the same variant and measure agreement on the top node (\textit{node@1}), top-3 set (\textit{node@3}, Jaccard), dominant family (\textit{family}), and best alternative (\textit{alt.}); this is the standard reproducibility-under-retraining notion, quantified in Table~\ref{tab:stability}.
    \item \textbf{Sanity}: \emph{learned saliency or architectural artefact?} The ratio of \textit{flip@1} between the trained policy and a randomised-weight copy, after~\cite{adebayo2018sanity}: a ratio near one means the attribution survives destroying the weights and cannot be trusted.
    \item \textbf{Actionability}: \emph{can a small instance change reach a different decision?} The counterfactual availability rate (any switch or make-feasible found) and the make-feasible rate.
\end{itemize}
The same grid applies to all (model, method) pairs, including the $\Att$ baseline. Because no prior joint encoder--decoder XAI protocol for neural CO solvers exists to benchmark against, the grid carries its own controls: the randomised-weight policy is the negative control for every attribution criterion (\S 5.5), and the $\Att$/$\EtoE$ cell (un-graph encoder, feasibility-by-construction decoder) is the reference against which the graph-bias and recourse effects are read.

\section{Encoder Results}


\begin{table}[t]
    \centering
    \small
    \resizebox{\columnwidth}{!}{\begin{tabular}{l rr rr rr}
    \toprule
    Model            & NMI         & macro-F1   & route-open F1 & flow-struct F1 & eff-rank   & silhouette \\
    \midrule
    \Att\ / \EtoE    &        0.48 &       0.23 &          0.48 &           0.94 &       14.4 &       0.36 \\
    \Att\ / \CA      &        0.50 &       0.22 &          0.50 &           0.94 &       14.2 &       0.39 \\
    \Unimp\ / \EtoE  &        0.57 &       0.36 &          1.00 &           0.93 &       15.7 &       0.19 \\
    \Unimp\ / \CA    &  \textbf{0.67} & \textbf{0.50} &     1.00 &           0.95 &       15.3 &       0.24 \\
    \Moe\ / \EtoE    &        0.55 &       0.36 &          1.00 &           0.91 & \textbf{15.95} & 0.20 \\
    \Moe\ / \CA      &        0.64 &       0.43 &          0.99 &  \textbf{0.97} &       15.5 &       0.22 \\
    \bottomrule
\end{tabular}
}
    \caption{Graph-level encoder summary. NMI / macro-F1 of $k$-means signatures vs.\ constraint families; F1 on two illustrative families; effective rank; best silhouette over $k$.}
    \label{tab:encoder}
\end{table}

\subsection{Predictability and Distributed Coding}

Table~\ref{tab:encoder} reports the supervised picture. Both graph encoders dominate $\Att$ on signature NMI ($0.55$--$0.67$ vs.\ $0.48$--$0.50$) and on multi-class macro-F1 ($0.36$--$0.50$ vs.\ $0.22$--$0.23$). They reach the ceiling on simple binary indicators (\emph{e.g.}\ route openness, $1.00$ vs.\ $0.48$--$0.50$). Pairing the graph encoders with the $\CA$ decoder adds another $0.05$--$0.14$ on NMI on top of either $\EtoE$ pairing, so the ranking is stable: $\Unimp/\CA \succ \Moe/\CA \succ$ either $\EtoE$ variant of the graph family $\succ \Att$.

The unsupervised PCA/ICA readouts add a more nuanced picture. Read on a single dominant axis, $\Moe$ underperforms $\Unimp$ on several constraint families. As soon as we allow a top-$k$ subspace, $\Moe$ catches up to within $\sim 0.04$ of $\Unimp$. The implication is that $\Moe$ does not encode \emph{less} information; it encodes it in a more distributed, non-axis-aligned way. Without the second readout, a researcher would conclude that $\Moe$ is a worse encoder than $\Unimp$; with it, the difference is one of representational geometry, not capacity.

\subsection{Spontaneous Organization and Richness}

\begin{figure}[t]
    \centering
    \includegraphics[width=\columnwidth]{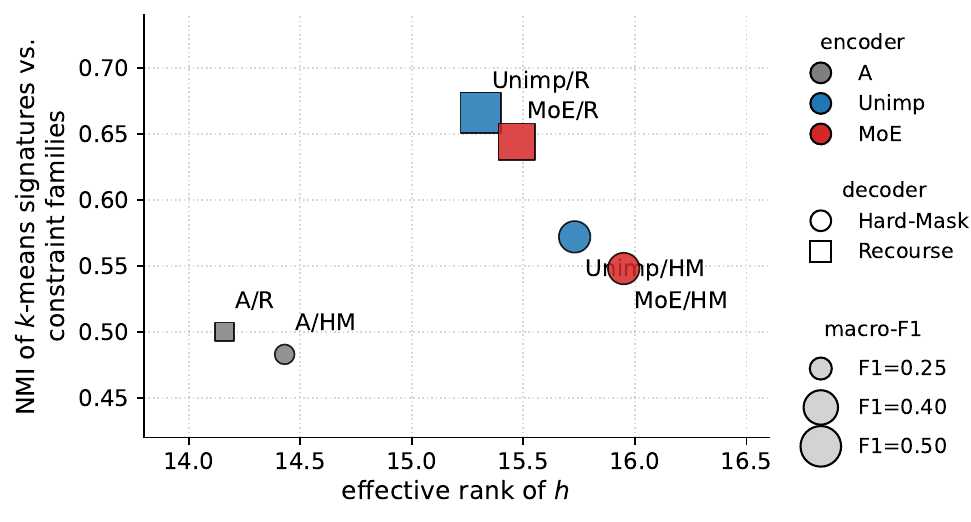}
    \caption{Joint reading of the encoder pillar. \emph{X}: effective rank of $h$. \emph{Y}: NMI of $k$-means signatures with constraint families. \emph{Point size}: multi-class macro-F1 of the constraint probe. Color encodes encoder family; marker encodes decoder variant. \Att\ sits at the bottom-left with small markers; $\Unimp$/\CA\ dominates on organization and F1; $\Moe$ is rich and distributed.}
    \label{fig:encoder}
\end{figure}

Even without supervision the latent organises by constraint family. NMI of $k$-means signatures (Table~\ref{tab:encoder}) ranks the encoders in the same order as the supervised probes, but the absolute values are higher than what the silhouette score alone would suggest, which is why combining both metrics is necessary. $\Att$ wins on silhouette ($0.36$--$0.39$, vs.\ $0.19$--$0.24$ for the graph encoders): its clusters are tight but smaller and less informative.

Effective rank and stable rank tell a complementary story. The graph encoders use a measurably wider portion of their representation space ($\Unimp$ $\sim 15.7$, $\Moe$ $\sim 15.95$) than $\Att$ ($\sim 14.4$), without collapse. Plotted jointly with NMI (Figure~\ref{fig:encoder}), the picture is unambiguous: $\Att$ at the bottom-left, $\Unimp$ leading on organization, $\Moe$ leading on richness while remaining well organised. A single-metric reading would have hidden the rank--organization trade-off.

\subsection{Concepts, Nodes, and Edges}

Moving from raw constraint indicators to constructed concepts (compactness, clustering, outliers, capacity pressure, combined tension) flattens the ranking: concept macro-F1 sits at $0.56$--$0.60$ across all six variants, $\Att$ slightly ahead on a few binary concepts. The graph advantage is thus on the raw constraint skeleton, not on derived concepts. The picture also changes with granularity. At the \emph{node} level the graph encoders still lead (node-role macro-F1 $0.62$--$0.64$ vs.\ $0.57$--$0.59$, widest on route position). At the \emph{edge} level it inverts on one informative axis: \emph{same-route} prediction is high across all encoders (AUC $0.90$--$0.92$), with $\Att$ on par or slightly above; attention encodes pairwise relational structure as well as a graph encoder does, even when its global summary is less organised. The graph advantage is therefore not uniform: it concentrates on global structure and per-node role, not on pairwise relational bookkeeping.

\subsection{Discovered Directions and Intervention}

\begin{figure*}[t]
    \centering
    \includegraphics[width=0.95\textwidth]{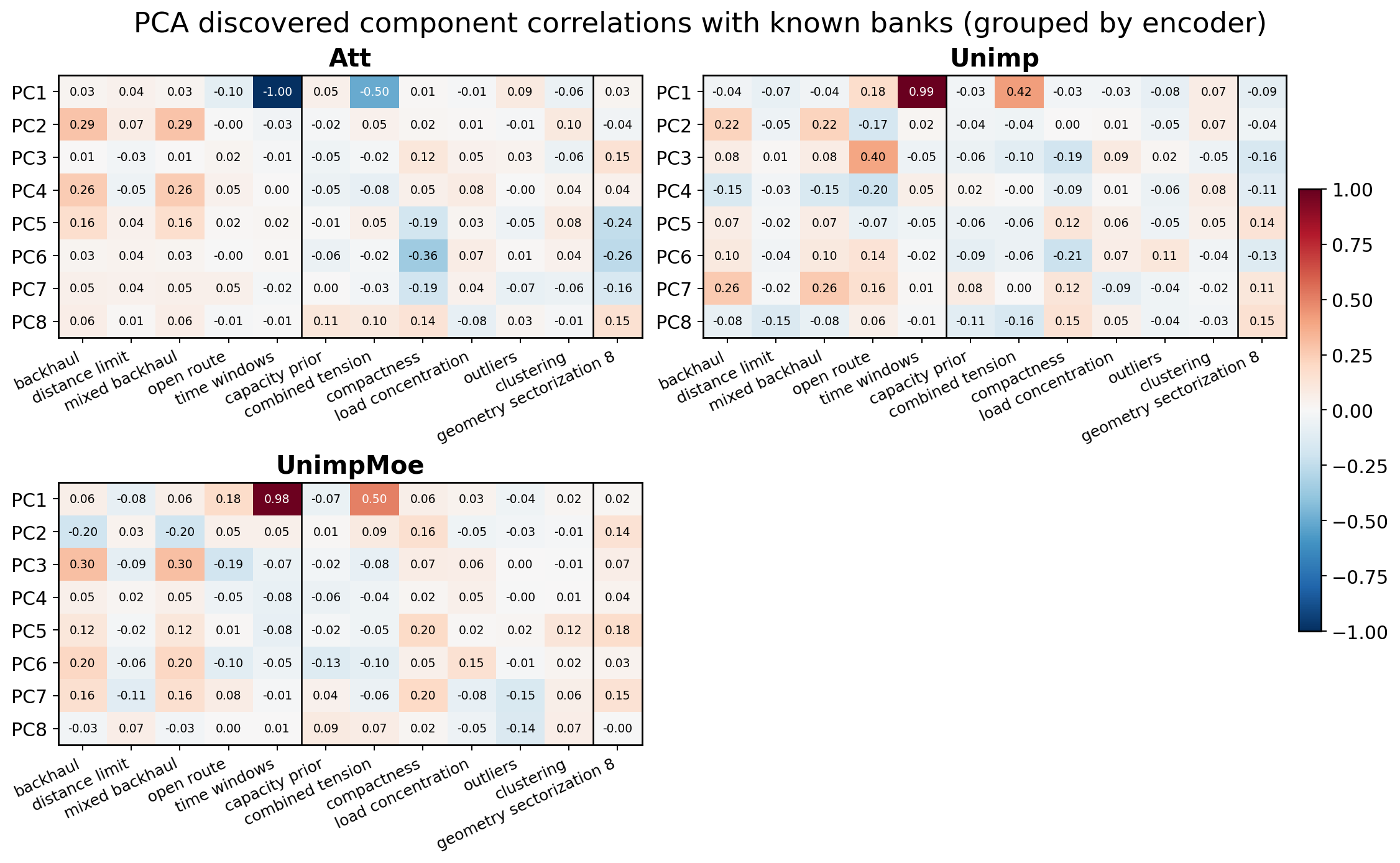}
    \caption{Alignment of the leading PCA components with the known constraint-and-concept bank, one panel per encoder family (each averages that family's two decoder variants). Rows are components PC1--PC8; columns are constraint indicators (left group) and derived concepts (right group); a cell is the signed correlation between the component and the reference, red positive, blue negative. Read the leading rows: \Att's PC1 locks onto route openness ($-1.00$) while the graph encoders' PC1 locks onto time windows ($\sim 0.98$); the remaining constraints spread across several components, the distributed pattern quantified in \S 4.1.}
    \label{fig:directions}
\end{figure*}

\begin{figure}[t]
    \centering
    \includegraphics[width=\columnwidth]{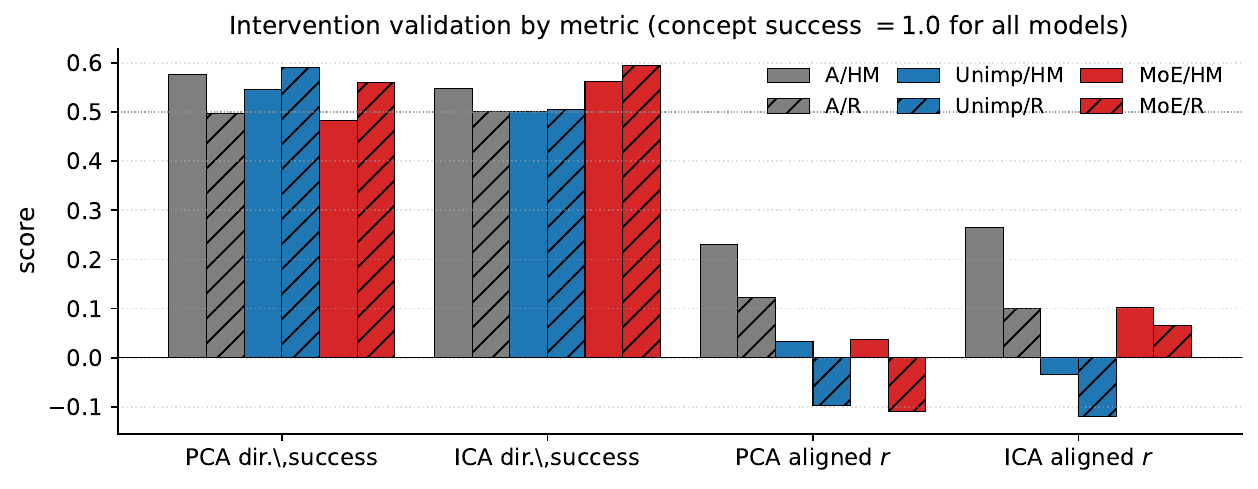}
    \caption{Intervention validation grouped by metric: directional success (PCA / ICA) and aligned-perturbation correlation (PCA / ICA). Bars within each metric compare the six (encoder $\times$ decoder) variants. Hatched bars are $\CA$ decoders; bar color is the encoder family. Concept-level success is $1.0$ for all variants and is omitted.}
    \label{fig:intervention}
\end{figure}

PCA and ICA decompositions of $h$ surface candidate latent axes. Aligning each axis against the known reference bank (Figure~\ref{fig:directions}), several principal components correlate strongly with one or two constraint families: route openness, geometry, and capacity pressure are recoverable from the leading axes, while time-window-related families remain spread over multiple components. This is qualitative evidence that the encoder's leading directions are interpretable without supervision, which we then test causally.

Intervention validation (Figure~\ref{fig:intervention}) pushes each aligned axis in the latent and checks whether the targeted concept moves in the predicted direction. Concept-level success is $1.00$ on all six variants, confirming that the alignments carry causal weight. Direction-level success is more modest at $0.48$--$0.59$, with PCA correlations small or negative for $\Unimp$ and $\Moe$, in line with the distributed-coding result of \S 4.1: the linear axis used for the intervention captures only part of the underlying signal in the GNN latents. Stability across seeds confirms that the leading axes themselves are reliable artefacts of the trained encoder, not of the random PCA/ICA initialisation.

\section{Decoder Results}

\subsection{Inference Cost: the Solvers Are Competitive}

\begin{table}[t]
    \centering
    \small
    \begin{tabular}{l rr}
    \toprule
    Method                 & mean gap        & best on    \\
    \midrule
    OR-Tools               & $1.94\%$        & 0          \\
    MTPOMO                 & $2.45\%$        & 0          \\
    MVMoE                  & $2.29\%$        & 0          \\
    CaDA                   & $1.71\%$        & 6          \\
    \midrule
    \Att\ / \EtoE          & $1.82\%$        & 0          \\
    \Unimp\ / \EtoE        & $1.54\%$        & 0          \\
    \Moe\ / \EtoE          & $\mathbf{1.39\%}$ & \textbf{10} \\
    \Att\ / \CA            & $2.87\%$        & 0          \\
    \Unimp\ / \CA          & $2.91\%$        & 0          \\
    \Moe\ / \CA            & $2.78\%$        & 0          \\
    \bottomrule
\end{tabular}

    \caption{Inference cost across 16 \MAVRP\ variants at $n=50$ (gap-to-BKS averaged, ``best on'' counts the variants where the method wins). $\EtoE$ checkpoints reach published-baseline level; $\Moe$/$\EtoE$ tops the grid. $\CA$ checkpoints pay an explicit $\sim 1\%$ gap for their interpretability properties.}
    \label{tab:inference}
\end{table}

Before reading the XAI signal, it matters that the policies under study are actually competitive. Table~\ref{tab:inference} reports the mean gap-to-BKS across $16$ \MAVRP\ variants at $n=50$, alongside the published learning baselines~\cite{berto2024rl4co}. The $\EtoE$ checkpoints land in the $1.39$--$1.82\%$ band, matching or beating every published baseline ($\Moe$/$\EtoE$ wins outright on $10$ of $16$ variants and CaDA on $6$). The $\CA$ checkpoints sit at $2.73$--$2.91\%$, roughly $1$ point worse than their $\EtoE$ counterparts and on the same order as MTPOMO/MVMoE. The cost is therefore an explicit trade-off: $\CA$ buys the interpretability properties documented below at a measurable but bounded gap penalty. This rules out the alternative reading in which $\CA$ ``looks more interpretable'' simply because its policy is degenerate.

\subsection{Constraint Families that Drive Decisions}

\begin{figure*}[t]
    \centering
    \includegraphics[width=0.8\textwidth]{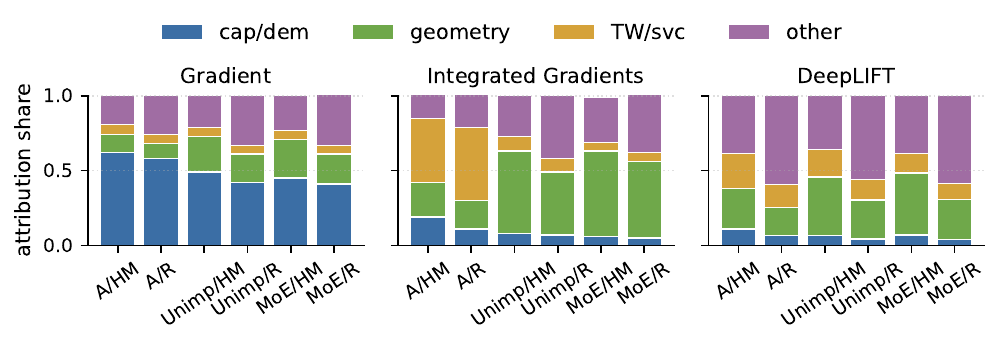}
    \caption{Stacked attribution share over the four constraint families, by (encoder $\times$ decoder), under gradient, integrated gradients, and DeepLIFT (left to right). The three methods agree on the $\CA$ decoder shifting mass into ``other'' but disagree on the dominant family, which is what motivates a multi-criterion grid.}
    \label{fig:share}
\end{figure*}

Figure~\ref{fig:share} reports the average constraint-family share of attribution mass. Two patterns emerge. First, the dominant family is \emph{method-sensitive}: gradient favours capacity/demand for $\Att$ ($\sim 60\%$) and geometry for $\Unimp$/$\Moe$; integrated gradients flips this to time-windows for $\Att$ ($>40\%$) and geometry for the graph encoders ($>50\%$); DeepLIFT pushes the largest share into the dynamic decoder-state ``other'' bucket, especially under $\CA$. The qualitative reading thus depends on the method; this is evidence for a multi-criterion grid, not a flaw of any single attribution. Second, the $\CA$ decoder inflates the ``other'' bucket (route-recourse and dynamic decoder-state features) by $\sim 7$--$11$ points under gradient/IG and $\sim 20$ under DeepLIFT: it shifts the decision horizon off raw inputs and onto its running state. The three methods therefore agree on the \emph{regime effect} even while disagreeing on the \emph{family ranking}.

\subsection{Contrastive: Recourse Exposes More Alternatives}

\begin{table}[t]
    \centering
    \small
    \resizebox{\columnwidth}{!}{\begin{tabular}{lrrr}
    \toprule
    Model                       & alt.\ rate & $\Delta$logit & $\Delta$logp \\
    \midrule
    \Att\ / \EtoE               &       0.74 &          2.01 &         9.92 \\
    \Att\ / \CA                 &       0.81 &          1.82 &         8.54 \\
    \Unimp\ / \EtoE             &       0.75 &          2.21 &         9.74 \\
    \Unimp\ / \CA               &  \bf 0.86  &          2.24 &         9.82 \\
    \Moe\ / \EtoE               &       0.76 &          1.99 &         9.21 \\
    \Moe\ / \CA                 &       0.83 &     \bf 2.29  &         9.55 \\
    \bottomrule
\end{tabular}
}
    \caption{Contrastive readout. ``alt.\ rate'' = fraction of steps with at least one feasible alternative; $\Delta$ is the chosen-vs-alternative margin.}
    \label{tab:contrastive}
\end{table}

Table~\ref{tab:contrastive} shows that the alternative-availability rate climbs from $0.74$--$0.76$ under the $\EtoE$ decoder to $0.81$--$0.86$ under the $\CA$ decoder, across all three encoders. The $\CA$ decoder systematically widens the local choice set without inflating the margin: gaps stay tight at $\sim 2$ logits, meaning many decisions are close calls, the setting in which contrastive explanations are most informative. Note that the contrastive readout is method-independent: it only requires knowing the chosen action and the next-best feasible candidate.

\subsection{Faithfulness}

\begin{table}[t]
    \centering
    \small
    \resizebox{\columnwidth}{!}{\begin{tabular}{l rrr rr}
    \toprule
    Model                       & flip@1 & flip@3 & flip@5 & $\Delta$logp@1 & $\Delta$logp@3 \\
    \midrule
    \Att\ / \EtoE               &  0.471 &  0.509 &  0.521 &           6.37 &           4.56 \\
    \Att\ / \CA                 &  0.474 &  0.499 &  0.518 &           6.00 &           3.62 \\
    \Unimp\ / \EtoE             &  0.468 &  0.474 &  0.474 &           7.28 &           5.29 \\
    \Unimp\ / \CA               & \bf 0.518 & \bf 0.537 & \bf 0.560 & \bf 8.01 & \bf 5.40 \\
    \Moe\ / \EtoE               &  0.463 &  0.493 &  0.510 &           7.15 &           5.27 \\
    \Moe\ / \CA                 &  0.487 &  0.497 &  0.527 &           7.41 &           4.93 \\
    \bottomrule
\end{tabular}
}
    \caption{Deletion faithfulness (gradient method, trained weights). $\Unimp$/\CA\ leads on every metric; the worst-best gap on flip@5 is $\sim 9$ points.}
    \label{tab:faith}
\end{table}

Table~\ref{tab:faith} reports the deletion-faithfulness numbers for the gradient method. Perturbing the single most-attributed node already flips the decision in $\sim 47$--$52\%$ of steps, and the log-prob drop is consistently larger for the graph encoders ($\Delta$logp@1 $7.15$--$8.01$) than for $\Att$ ($6.00$--$6.37$). $\Unimp$/\CA\ leads on every metric. The choice of attribution method barely moves faithfulness: averaged over all six variants, gradient, integrated gradients, and DeepLIFT stay within $0.02$ of each other on flip@1 and within $0.20$ on $\Delta$logp@3. They are equally faithful even though they disagree on which family they highlight (Figure~\ref{fig:share}): equal-fidelity attributions can tell different qualitative stories, which is exactly why the scorecard separates fidelity from concentration.

\subsection{Sanity}

\begin{table}[t]
    \centering
    \small
    \resizebox{\columnwidth}{!}{\begin{tabular}{l rr r rr}
    \toprule
    & \multicolumn{2}{c}{flip@1} & ratio & \multicolumn{2}{c}{$\Delta$logp@1} \\
    \cmidrule(lr){2-3}\cmidrule(lr){5-6}
    Encoder    & trained & random &  trained/rand    & trained & random \\
    \midrule
    \Att       &   0.47  &   0.27 &  $\sim 1.7\times$ &   6.18  &   0.01 \\
    \Unimp     &   0.49  &   0.12 &  $\sim 4.2\times$ &   7.65  &  -0.32 \\
    \Moe       &   0.47  &   0.14 &  $\sim 3.5\times$ &   7.28  &  -0.40 \\
    \bottomrule
\end{tabular}
}
    \caption{Sanity check: trained vs.\ randomized-weight policies (gradient method). Graph encoders pass much more decisively than \Att.}
    \label{tab:sanity}
\end{table}

Table~\ref{tab:sanity} contrasts trained and randomized weights. All encoders pass: trained policies flip more often and the log-prob drops are an order of magnitude larger. Strikingly, the trained/random ratio on flip@1 is $\sim 1.7\times$ for $\Att$ but $\sim 3.5$--$4.2\times$ for the graph encoders. Part of $\Att$'s saliency is therefore architectural rather than learned, a finding only the sanity criterion exposes, and one which would silently bias a head-to-head comparison restricted to flip@$k$ alone.

\subsection{Stability}

\begin{table}[t]
    \centering
    \small
    \resizebox{\columnwidth}{!}{\begin{tabular}{l rr rr}
    \toprule
    Model                       & node@1 & node@3 & family & alt.\ \\
    \midrule
    \Att\ / \EtoE               &  0.504 &  0.514 &  0.985 &  0.509 \\
    \Att\ / \CA                 &  0.512 &  0.519 &  0.948 &  0.508 \\
    \Unimp\ / \EtoE             &  0.517 &  0.512 &  0.808 &  0.506 \\
    \Unimp\ / \CA               &  0.520 &  0.518 &  0.627 &  0.515 \\
    \Moe\ / \EtoE               &  0.509 &  0.521 &  0.894 &  0.509 \\
    \Moe\ / \CA                 &  0.504 &  0.516 &  0.631 &  0.505 \\
    \bottomrule
\end{tabular}
}
    \caption{Cross-seed stability (gradient method): mean agreement of explanations across independently seeded checkpoints of the same variant, over all $\binom{4}{2}\!=\!6$ seed pairs. \textit{node@1} / \textit{node@3}: agreement on the top attributed node / top-3 node set (Jaccard); \textit{family}: agreement on the dominant constraint family; \textit{alt.}: agreement on the identified best alternative. Higher is more reproducible.}
    \label{tab:stability}
\end{table}

Stability is the pillar most easily asserted and least often measured, so we quantify it directly. Table~\ref{tab:stability} re-runs the gradient attribution on four independently seeded checkpoints of each variant and reports how often the explanations agree across the six seed pairs. Node-level saliency reproduces only moderately and uniformly: the top node agrees about half the time (\textit{node@1} $0.50$--$0.52$), the top-3 set overlaps similarly (\textit{node@3} $0.51$--$0.52$), and the named best alternative shows the same near-tie (\textit{alt.} $\sim 0.51$). This floor is expected: many nodes carry comparable mass and the margins are tight ($\sim 2$ logits, \S 5.3), so the exact ranking is seed-sensitive even when the story is not.

The dominant \emph{family} is far more reproducible, and how reproducible is itself informative. $\Att$ and the $\EtoE$ policies pin the same family almost every time ($0.95$--$0.99$ for $\Att$; $0.81$--$0.89$ for $\Unimp$/$\Moe$ under $\EtoE$), whereas the graph encoders under $\CA$ fall to $\sim 0.63$. This mirrors the constraint-share result (\S 5.2): when a policy spreads its decision onto distributed decoder-state features, \emph{which} family reads as dominant is less determined and flips more across seeds. The property that makes $\CA$ explanations richer thus makes their one-line family summary less stable, a trade-off a fidelity-only reading would hide. The pattern is method-robust: IG and DeepLIFT give the same $\sim 0.50$ node floor and family agreement in the $0.63$--$0.85$ band.

\subsection{Counterfactuals and Make-Feasible}

The search finds at least one switch or make-feasible candidate at $32$--$49\%$ of steps, with $\Unimp$/\CA\ on top ($0.49$). The \emph{make-feasible} mode has rate $0.00$ for all three $\EtoE$ decoders and $0.03$--$0.06$ for all three $\CA$ decoders, and $\CA$'s perturbations are smaller ($0.82$--$1.23$ vs.\ $1.03$--$1.46$): smaller and more frequent. By itself the asymmetry is mechanical ($\EtoE$ masks infeasible candidates out by construction, so the search has nothing to operate on), which raises the real question: could the $\EtoE$ \emph{policy} produce make-feasible counterfactuals if we lifted that restriction? We test this next.

\paragraph{Representational ablation.} To isolate the policy effect from the mask effect, we re-ran the search on the three $\EtoE$ checkpoints with the alt-selector externally widened to the visited-only candidate pool, the same pool $\CA$ sees at training. The widened alt-availability rate matches the $\CA$ regime ($0.88$, $0.72$, $0.86$ for $\Att$, $\Unimp$, $\Moe$), yet the make-feasible rate \emph{stays at $0.00$} across all three encoders. The $\EtoE$ policy was never trained to assign meaningful logit mass to infeasible candidates, so the next-best alternative the alt-selector returns is almost always already feasible and the perturbation has nothing to re-feasibilize. The $\EtoE$ vs.\ $\CA$ make-feasible gap is therefore a property of the trained policy, not of the mask: training under recourse cost is what produces a policy whose counterfactual neighbourhood contains actionable make-feasible moves.

\subsection{Scorecard}

\begin{figure}[t]
    \centering
    \includegraphics[width=\columnwidth]{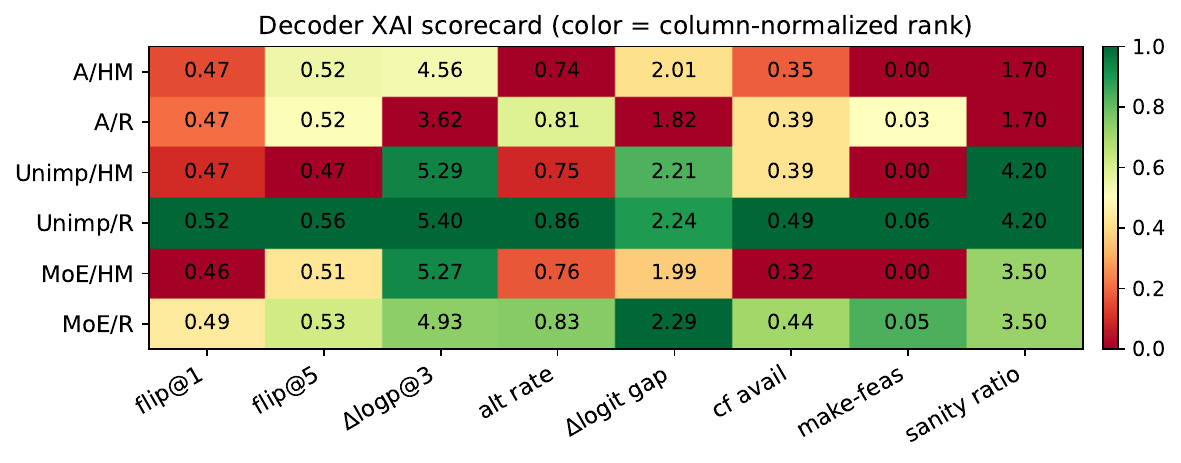}
    \caption{Decoder XAI scorecard across eight metrics covering fidelity, contrast, actionability, and sanity. Color is column-normalized rank; cell text is the raw value.}
    \label{fig:scorecard}
\end{figure}

Figure~\ref{fig:scorecard} folds the eight headline metrics into one heatmap. It covers four of the five criteria (fidelity, concentration, sanity, actionability) as per-checkpoint columns; the fifth, stability, is a cross-seed quantity reported separately in Table~\ref{tab:stability}. $\Unimp$/\CA\ wins on five of the eight metrics, including faithfulness and counterfactual availability, and $\Att$/\EtoE\ wins on none; the $\CA$ rows are uniformly greener than the $\EtoE$ rows, except on $\Delta$logp@3. Read with the stability table the picture is nuanced rather than a clean sweep: $\CA$ buys richer, more actionable explanations at the cost of a less stable single-family summary. The scorecard makes both the encoder ranking and the decoder effect visible at a glance, which is what an operations team needs from a comparability tool.

\section{Related Work}

Linear probing of frozen representations~\cite{alain2016understanding} is standard for linear decodability; we extend it to structural MAVRP constraints and add axis-unique vs.\ subspace readouts to expose distributed coding. Gradient, integrated gradients~\cite{sundararajan2017axiomatic}, and DeepLIFT~\cite{shrikumar2017learning} via Captum~\cite{kokhlikyan2020captum} are mature in vision/NLP but rarely combined on neural CO solvers; we adopt the sanity protocol of~\cite{adebayo2018sanity} as the criterion that flags architecturally-induced saliency. GNNExplainer~\cite{ying2019gnnexplainer} and attention-flow methods~\cite{abnar2020quantifying} target a single forward pass on a static graph, whereas we attribute autoregressive \emph{actions} whose state evolves with each decision, forcing the per-step contrastive reading. Classical counterfactuals seek the smallest input change that flips a class; the \emph{make-feasible} mode (smallest change that turns an infeasible alternative feasible) is, to our knowledge, specific to recourse-style policies with no direct analogue in classification XAI.

\section{Discussion and Conclusion}

\paragraph{What XAI revealed.} Architecture matters for both representation and decision. Graph inductive bias lifts probe quality, sharpens spontaneous organization, and passes the sanity check more decisively; and the MoE encoder is not less informative than $\Unimp$ but represents constraints in a distributed subspace, visible only beyond axis-aligned probes. None of this could be read off the inference cost alone.

\paragraph{Decoder matters.} The $\CA$ decoder widens the contrastive landscape and shifts attribution toward dynamic decoder-state features; crucially, its \emph{training regime} is what produces a policy whose neighbourhood contains make-feasible counterfactuals, which the $\EtoE$ policy fails to produce even when fed infeasible alternatives externally. The decoder is therefore not just a performance choice but an \emph{interpretability} choice: it changes what kind of explanation the solver can produce at all.

\paragraph{Scale robustness.} On all six $n=100$ checkpoints the gradient-method picture holds: graph encoders lead $\Att$ on faithfulness ($\Delta$logp@1 up to $10.7$ vs.\ $8.9$--$9.5$), $\CA$ widens the contrastive landscape (alt rate $0.89$--$0.91$ vs.\ $0.84$--$0.85$), and the make-feasible asymmetry is textbook ($0.03$ for every $\CA$, $0.00$ for every $\EtoE$). The protocol is not overfit to $n=50$.

\paragraph{Limits and outlook.} Three limits scope the claims: the encoder pillar relies on a \emph{supervised} concept dictionary, so it tests only for structure we thought to name; the counterfactual search is a first-order, fixed-budget proposal (\S 3.2), so its rates are lower bounds a heavier optimiser would raise; and the $n=100$ check uses one seed per cell, so only the $n=50$ grid is variance-controlled. Four directions follow: (i) \emph{dispatcher-in-the-loop} evaluation in the spirit of HCXAI, measuring whether operators diagnose, accept, or override routes faster with the explanations in hand; (ii) a stability-regularised or multi-family attribution that resolves the stability--richness trade-off we surface (\S 5.4); (iii) porting the make-feasible counterfactual to other recourse-trained policies (scheduling, bin packing, soft-constrained dispatching) to test how VRP-specific the effect is; and (iv) growing this grid and its controls into the standard XAI benchmark neural CO still lacks. The protocol is reproducible on any \MAVRP\ checkpoint trained with the rl4co interface~\cite{berto2024rl4co}.

\bibliographystyle{named}
\bibliography{references}

\end{document}